\documentclass[11pt]{article}
\usepackage{tabularx}
\usepackage{wrapfig}
\usepackage[table]{xcolor}
\usepackage{booktabs}
\usepackage{tcolorbox}
\usepackage{enumitem}
\usepackage{graphicx}
\usepackage{float}
\usepackage{makecell}
\usepackage{caption}
\usepackage[table]{xcolor}
\usepackage{xcolor}
\usepackage{multirow}
\usepackage[final]{acl}
\usepackage{listings}
\usepackage{subcaption}
\usepackage{placeins}

\title{A French OSCE Dialogue Dataset and Controllable Virtual Patient System for Clinical Training}

\author{Doria Bonzi$^1$, Tom Bourgeade$^1$, Fabrice Lefèvre$^2$, Irina Illina$^1$ \\
        $^1$Lorraine Université, CNRS, Inria, LORIA, Nancy, France \\
        $^2$Avignon Université, LIA, UPR 4128, Avignon, France \\
        \texttt{doria.bonzi@loria.fr, tom.bourgeade@loria.fr, irina.illina@loria.fr} \\
        \texttt{fabrice.lefevre@univ-avignon.fr}}

\begin{document}

\maketitle

\begin{abstract}
The clinical and communication skills of medical students are commonly assessed through Objective Structured Clinical Examinations (OSCEs), which consist of brief scenario-driven simulations of doctor-patient interactions. However, training is often limited by the low availability of human standardized patients, motivating the development of realistic virtual patients (VPs).
To address this gap, we introduce a French OSCE dialogue dataset comprising 240 student–patient training interactions. We build upon it a controllable LLM-based pipeline to generate synthetic OSCE dialogues. The pipeline integrates modular components, such as retrieval-based grounding and a reflection loop, to ensure patient fidelity, coherence, and realism. 
Additionally, we propose a multi-level evaluation framework assessing patient simulation quality, student performance, and linguistic quality, using an LLM-as-a-Judge approach.
Experiments suggest that controllability modules generally improve patient fidelity and student evaluation consistency.
Finally, we also implement an interactive prototype in which students can practice with a VP and receive automatic feedback.
\end{abstract}

\section{Introduction}

Objective Structured Clinical Examinations (OSCEs) are used to assess both clinical reasoning and communication skills in medical education. In France, these exams involve medical students playing the role of a physician in 7 to 10-minute scenario-based simulated interactions with %either 
a \textbf{standardized patient} (SP), %or \textbf{standardized health professional} (SHP)
under the observation of an \textbf{evaluator}. A SP is defined as a trained individual who portrays a predefined patient scenario for the training and assessment of healthcare students \cite{Lewis2017ASPE}. Each OSCE scenario is referred to as a ``\textbf{station},'' and can cover various medical interactions, ranging from patient history taking and analyzing results to breaking bad news. By design, OSCEs approximate real clinical encounters and intentionally simplify many aspects for pedagogical purposes. Communication skills play a central role in OSCE performance; however, student training is limited by the availability of human standardized patients and evaluators, leading to scalability and cost issues for repeated practice.

\begin{figure*}[t]
\centering
\includegraphics[width=0.8\linewidth]{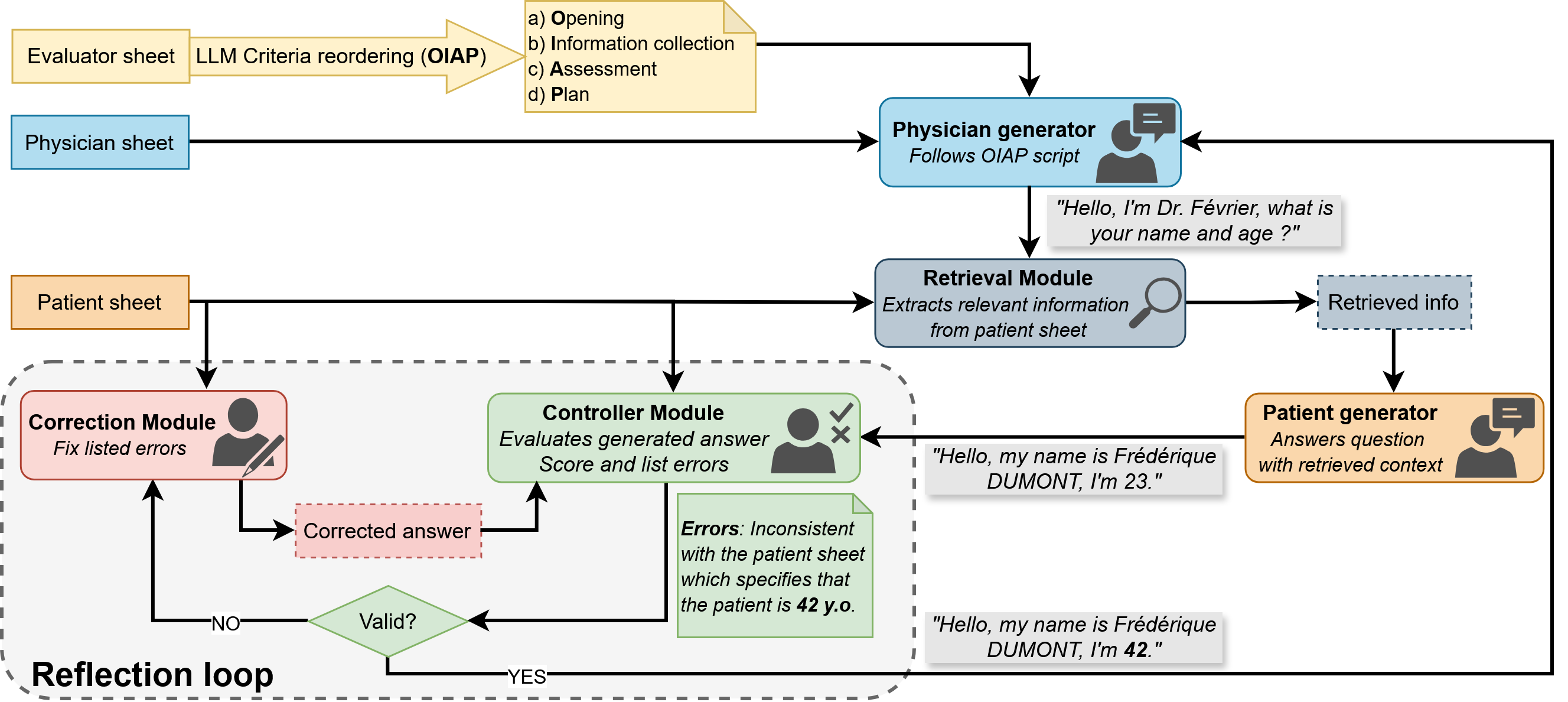}
\caption{Overview of the controllable LLM-based dialogue generation pipeline for OSCE simulations in automatic mode. The physician generator follows a criteria-based script; the patient generator can be augmented by a retrieval module, and the reflection loop with controller-corrector modules can be activated to control patient faithfulness through a verification-correction loop.} %In interactive mode (not shown), the physician agent is replaced by a user (student) who receives the physician sheet and conducts the consultation with the VP.}
\label{fig:system_overview}
\end{figure*}

\textbf{Virtual Patient} (\textbf{VP}) systems have been proposed to address these problems, relying on rule-based or script-driven approaches %, which ensured controlled interactions but often resulted in limited adaptability to learner input 
\cite{elzini2019,CampillosLlanos2019VirtualPatientDialogue,laleye_SemanticSimilarityImprove_2020}. Data-driven and LLM-based VP systems have demonstrated improved fluency and realism \cite{voigt2025llmpoweredvirtualpatientagents, GarciaTorres2024VirtualPatients, cook2025virtual}. Some studies explored automated feedback for OSCE assessments using LLM-as-a-Judge \cite{shakur2024largelanguagemodelsmedical, campbell2025aiosce, huang2026osce}.
Despite these advances, current LLM-based VP systems often lack controllability, hindering consistent adherence to predefined clinical stations. Recent work has also highlighted the lack of standardized and reproducible evaluation frameworks in LLM-based VP systems \cite{li2026llmvp}, essential for reliable assessment. 

Publicly available OSCE-related datasets are rare, especially in French. Existing OSCE datasets in English focus on specific clinical tasks or domains \cite{Fareez2022SimulatedPatientInterviews, saley_MediTODEnglishDialogue_2024}. Large-scale medical dialogue resources \cite{liu-etal-2024-meddialog,mts-dialog} are generally not aligned with OSCE constraints or lack the interactional and evaluative structure required for examination settings. French resources remain limited in size and scope %, and frequently mix heterogeneous interaction types or partially synthetic data
\cite{laleye_french_2020}. This lack of French OSCE data restricts the development of data-driven training and evaluation tools.

To address these challenges, we introduce in this paper:
\begin{enumerate}[label=(\arabic*),itemsep=0pt,parsep=1pt,leftmargin=0pt,itemindent=*,topsep=0pt]

\item a dataset of 240 recorded and transcribed French OSCE training dialogues available on Zenodo\footnote{Dataset may be accessed upon request: \href{https://zenodo.org/records/20719833}{zenodo.org/records/20719833}};%, along with 192 OSCE stations organized under a dedicated taxonomy;
\item a controllable LLM-based OSCE dialogue generation pipeline with modular components, supporting both automatic and interactive modes;
\item a multi-level LLM-as-a-Judge evaluation framework assessing generated dialogues and recorded OSCE interactions.

\end{enumerate}

\section{Related work}

\noindent\textbf{Virtual patients and LLM-based simulations:} Recent work on LLM-based VPs focuses on realism, interactivity, automatic scoring and feedback \cite{voigt2025llmpoweredvirtualpatientagents,GarciaTorres2024VirtualPatients} and patient data fidelity \cite{wang-etal-2024-patient,laverde2025virtualpatient}. Embodied VP have been explored as simulation-based tools, emphasizing realistic behavioral interactions \cite{chaby2022embodied}. Agentic approaches used structured patient data combined with multi-agent RAG workflows to enable controllability \cite{yu_SimulatedPatientSystems_2025}. However, these works were rarely anchored in real OSCE recordings, limiting their grounding in authentic interactions. 

\noindent\textbf{Medical dialogue and synthetic dialogue generation:}

In the context of OSCEs, \citet{Fareez2022SimulatedPatientInterviews} introduced a dataset of simulated patient interviews focused on respiratory cases. 
Building on this work, \citet{saley_MediTODEnglishDialogue_2024} released an English dataset for medical history-taking in OSCE format. French medical dialogue resources remain limited. \citet{laleye_french_2020} introduced a small annotated French corpus combining generated dialogues and interactions between medical students and patients. %While this dataset includes consultation-style interactions, it is limited in size and scope and is not explicitly aligned with OSCE evaluation constraints.
\citet{chen2023llmpsychiatry} has investigated LLM-based dual-agent simulations, emulating both physicians and patients for clinical dialogue generation.

Large-scale medical dialogue datasets provide broad coverage of clinical interactions but are not designed for OSCE-specific scenarios \cite{Jepson2017OneInAMillion,zeng_MedDialogLargescaleMedical_2020, liu-etal-2024-meddialog,mts-dialog}. In parallel, LLM-based dialogue generation approaches improved fluency and realism but lack strict controllability and alignment with structured exam settings \cite{das2024syntheticpatientphysiciandialoguegeneration,Wang_2024}. 

\noindent\textbf{Reflective prompting, and LLM-based evaluation:} Recent advances in LLM prompting have introduced self-reflection and critique mechanisms to improve factuality, coherence, and task adherence in complex generation tasks \cite{agrawal2026gepa,chirkova_LLMasaqualitativejudgeAutomatingError_2026,li_ZeroShotLanguageAgent_2023}. In parallel, LLM-as-a-Judge paradigms have emerged to evaluate conversational agents at scale across multiple criteria, including dialogue quality and OSCE performance \cite{shakur2024largelanguagemodelsmedical, campbell2025aiosce}. \citet{gu_SurveyLLMasajudge_2026} highlighted their growing adoption, with applications ranging from human-machine dialogue assessment \cite{njifenjou-etal-2025-enabling, njifenjou2024roleplayzeroshotpromptinglarge} to OSCE scenarios \cite{shakur2024largelanguagemodelsmedical, campbell2025aiosce} and benchmarking conversational agents in controlled settings \cite{zheng_JudgingLLMasaJudgeMTBench_2023}. These approaches motivate the design of structured evaluation and reflective stages, which we incorporate in our pipeline to assess and iteratively improve our generated OSCE dialogues.\\

\noindent Building on prior work, we distinguish our approach in three ways: (1) a focus on French OSCE stations, addressing the scarcity of French medical dialogue resources; (2) a modular, controllable LLM dialogue generation pipeline integrating information retrieval and self-reflection, grounded in OSCE data across more than ten medical specialties; (3) a multi-faceted evaluation covering linguistic and clinical performance.

\section{Proposed French OSCE dialogue dataset}% overview}

\begin{figure*}[t]
\centering
\begin{subfigure}[b]{.40\textwidth}
\small
\begin{tabular}{lc}
\hline
\textbf{Component} & \textbf{Count} \\%& \textbf{Description} \\
\hline
Total OSCE stations & 192 \\%& Fictional clinical cases used for recorded and generated data \\
\hline
\multicolumn{2}{l}{\textbf{Recorded dialogues}} \\
OSCE stations recorded & 23 \\%& With at least one recorded interaction \\
Recorded dialogues & 240 \\%& Student-patient interactions recorded during OSCE training sessions, automatically diarized and transcribed \\
%\hspace{1em} with video & 85 & \\
%Transcribed dialogues & 240 & Automatically diarized and transcribed with speaker labels \\
%Corrected transcriptions & 26 & Manually reviewed and corrected by human annotators \\
Total audio duration (hours) & 30 \\%& Cumulative duration of all recordings \\
\hline
\multicolumn{2}{l}{\textbf{Generated dialogues}} \\
OSCE stations used & 11 \\%& Cases used for dialogue generation \\
Generated dialogues & 792 \\%& Dialogues generated with our proposed pipeline \\
Total generated words & 1,22M \\
\hline
\end{tabular}
\caption{Dataset statistics: recorded and generated dialogues.}
\label{tab:dataset_structure}
\end{subfigure}
\hspace{20pt}
%\begin{subtable}[b]{.35\textwidth}
%\small
%\setlength{\tabcolsep}{4pt}
%\renewcommand{\arraystretch}{1.1}
%\begin{tabular}{lcc}
%\toprule
%\textbf{Interlocutor} & \textbf{No doc.} & \textbf{With doc.} \\
%\midrule
%Patient      & \cellcolor{blue!70}\textcolor{white}{89} & \cellcolor{blue!30}35 \\
%Professional & \cellcolor{blue!15}18 & \cellcolor{blue!20}15 \\
%Relative     & \cellcolor{blue!25}27 & \cellcolor{blue!10}8 \\
%\bottomrule
%\end{tabular}
%\caption{Taxonomy and number of OSCE stations per type. %defined with interlocutor type and document-based reasoning. 
%Each cell indicates the number of stations in the corresponding category.
%}
%\label{tab:table1}
%\end{subtable}
\begin{subfigure}[b]{.35\textwidth}
\small
\setlength{\tabcolsep}{4pt}
\renewcommand{\arraystretch}{1.1}
\centering
\includegraphics[width=1.01\linewidth]{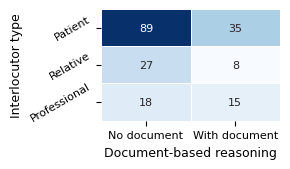}
%\caption{Taxonomy of OSCE stations defined along two factors: interlocutor type and document-based reasoning. Each cell indicates the number of stations in the corresponding category.}\label{fig:table1}
\caption{OSCE station classification and number of stations per category.} %defined with interlocutor type and document-based reasoning. 
%Each cell indicates the number of stations in the corresponding category.
\label{tab:table1}
\end{subfigure}
\caption{Overview of the French OSCE dialogue dataset, both recorded and generated.}
\label{tab:table1_global}
\end{figure*}

Our proposed dataset consists of: (i) a corpus of 240 recorded OSCE training dialogues, representing a total of 30 hours of audio, and (ii) a corpus of 792 generated dialogues produced using different experimental configurations of our controllable LLM-based generation pipeline (see~\autoref{tab:dataset_structure}).

\subsection{Data collection: OSCE stations and recorded dialogues}

\paragraph{OSCE stations:}
A total of 192 OSCE clinical stations were collected for this dataset across more than 10 medical specialties. These stations are authored by medical teachers and healthcare professionals. Each OSCE station consists of three complementary documents made available through a dedicated online platform: a \textbf{physician sheet} addressed to the evaluated student, a \textbf{patient sheet}, and an \textbf{evaluator sheet}.

Each OSCE station was mapped to a category from a dialogue-generation-focused \textbf{difficulty classification} developed specifically for this work, with: (i) the type of interlocutor and (ii) the requirement for document analysis (\autoref{tab:table1}). 

\paragraph{OSCE recorded dialogues:} \label{sec:OSCE_recordings}
We recorded 240 played physician-patient dialogues involving 99 sixth-year medical students across 23 distinct OSCE stations. These recordings were collected during OSCE training sessions organized weekly by the local student association, which are designed to closely replicate the conditions of the national OSCE examinations in France. Due to the limited availability of external volunteers, all roles (physician, patient, and evaluator) were performed by students. Although this is not ideal, it was an unavoidable constraint in our setting. As participants are neither professional actors nor trained OSCE evaluators, this setup may affect the realism of the interactions and evaluations, and may partially account for differences observed when comparing these dialogues with synthetically generated ones. 

Each OSCE training session lasted 8 minutes, followed by a 2-minute feedback period from the evaluator. All participants provided written informed consent for the use of their recordings for research purposes. 
Audio was captured using wireless clip-on microphones. All recordings were acquired using the same software, producing synchronized two-channel WAV files. More details on the recording protocol are available in Appendix~\ref{appendix:recording}. 

\paragraph{Recorded stations annotations:} Each recorded station was manually given labels with: one of 12 \textbf{specialties} (e.g., neurology, pediatrics, gastroenterology), a \textbf{consultation type} (e.g., emergency, follow-up), and one or more \textbf{objectives} (e.g., diagnosis, breaking bad news, history-taking, patient education). 

All annotations were performed by a single annotator through careful examination of the OSCE station materials, particularly the physician and evaluator sheets, which typically provide the station context, including specialty, consultation type, and primary objectives (see Appendix~\ref{appendix:case_annotations}). 

\paragraph{Automatic transcription:} \label{sec:transcription} All recorded dialogues were automatically diarized using the \texttt{precision-2} model \cite{pyannote_precision2}, and transcribed using the \texttt{faster-whisper-large-v3-turbo} model \cite{whisper_large_v3_turbo,radford_RobustSpeechRecognition_2023a}. 

\paragraph{Transcription evaluation:} To assess transcription quality, Word Error Rate (WER) was computed by comparing automatic transcriptions with manually corrected versions. A subset of 26 automatically transcribed dialogues, representing approximately 11\% of the recorded corpus, was manually corrected by 15 French-fluent annotators using the original audio recordings. %\textcolor{blue}{This correction-based protocol was chosen for practical reasons, as it substantially reduced annotation time compared to producing fully independent transcriptions from scratch.}% and, when available, the corresponding video data. Each annotator corrected between one and three recordings. %Manual correction addressed both transcription errors and diarization errors. 

The resulting WER of 7.7\% suggests good transcription quality for real-world clinical dialogues, though computed on a limited sample. More information on WER computation is available in Appendix~\ref{appendix:wer}.
Recurrent errors included names, acronyms (e.g., \textit{SMUR, ECOS}), domain-specific medical terminology (e.g., \textit{dyspnée}; ``shortness of breath''), occasional diarization errors due to cross-talk and disfluencies such as hesitations or repetitions.%, which were outside the scope of this work.

% Dialogue generation pipeline
\section{Controlled dialogue generation pipeline} \label{sec:proposed_method}

To benchmark LLM-based virtual patients against student-acted patients in OSCE training sessions, we synthetically generated 792 dialogues from structured OSCE station sheets. We focused on 11 stations for which real training recordings were available, excluding multimodal stations involving document analysis (e.g., radiology images). In this work, we deliberately restrict the problem to text-based interactions, excluding speech and multimodal information. %Multimodal stations involving document-based analysis (e.g., radiology images) were excluded.

\begin{figure}[t]
\centering
\includegraphics[width=1\linewidth]{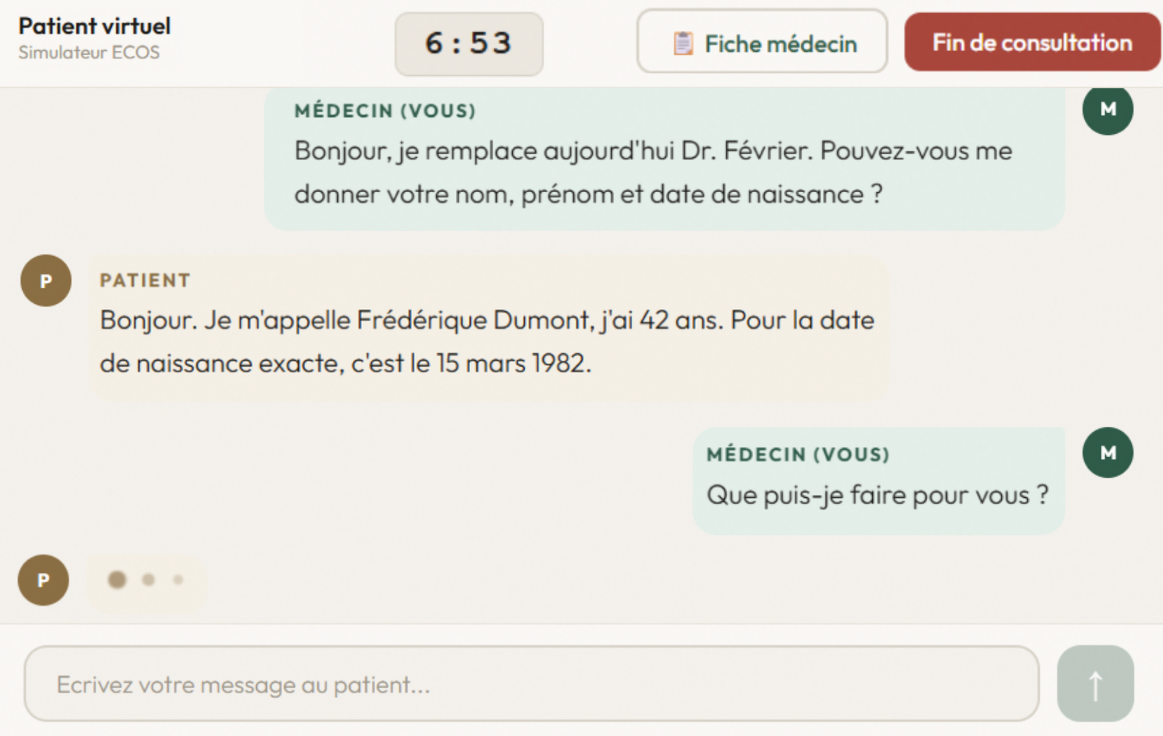}
\caption{Interactive demo of our VP system. The student interacts with the VP via messages and can consult the physician sheet at any time. An 8-minute timer simulates OSCE conditions.} 
\label{fig:demo}
\end{figure}

Our proposed approach (\autoref{fig:system_overview}) frames dialogue generation around \textbf{controllability}, aiming to ensure fidelity to patient profiles, coherence, and realism. The pipeline incorporates multiple levels of control with \textbf{retrieval-based grounding} and consistency enforcement through a \textbf{reflection loop}.
The proposed system generates VP utterances interacting with a simulated %student playing the role of (and hereafter referred to as) the 
\textbf{physician} in an OSCE setting, constrained to a fixed consultation duration of 8 minutes. The architecture is modular and LLM-based and supports two operating modes:
\begin{enumerate}[itemsep=1pt,parsep=1pt,topsep=1pt,itemindent=0pt,leftmargin=1.0em]
\item an \textbf{interactive mode}, where a student can train as the physician, via our prototype interface (see \autoref{fig:demo}). At the end of the interaction, students receive an automatic LLM-based evaluation of their performance, including feedback on addressed criteria and overall communication quality (see~\ref{sec:student_perf}). %This mode is currently a text-based prototype; integration of voice-based interaction is planned as future work.
\item an \textbf{automatic mode}, where both physician and patient are LLM-generated, enabling large-scale dialogue dataset generation.
\end{enumerate}
In both modes, VP utterances are controlled to ensure fidelity and consistency with the patient profile. The system relies on up to four distinct LLM instances assigned to specific roles: \textbf{retrieval}, \textbf{generation}, \textbf{control}, and \textbf{correction}. This modular design enables independent model selection and parameter tuning for each component.
Dialogue generation is grounded in three components:
\begin{enumerate}[itemsep=1pt,parsep=1pt,topsep=1pt,itemindent=0pt,leftmargin=1.5em]
\item \textbf{Patient sheet}, describing the simulated patient’s identity, medical history, symptoms, and behavior (e.g., \textit{Frédérique Dumont} consulting for an extension of medical leave);
\item \textbf{Physician sheet}, providing the clinical context and setting, available informations, and consultation objectives (e.g., current medication and information-gathering goals);
\item \textbf{Evaluation sheet}, listing OSCE criteria the physician should address, each corresponding to a specific clinical action or information (e.g., assessing treatment effectiveness).
\end{enumerate}

\paragraph{Dialogue script and criteria reordering:} \label{sec:oiap} In automatic mode, dialogue generation is guided by a script derived from the evaluation checklist of the OSCE station. Following prior work \cite{huang2026osce}, criteria are reordered into dialogue phases: \textbf{O}pening and preparation, \textbf{I}nformation collection, \textbf{A}ssessment, and \textbf{P}lan and conclusion (OIAP), inspired by SOAP \cite{podder2026soap}. This reordering is performed by an LLM before generation and used to guide physician utterances step by step, acting as a dialogue script \cite{huang2026osce}. 
We define two generation modes: a \textbf{standard} mode following the OIAP structure, and a \textbf{random} mode where intermediate phases and criteria order are shuffled while preserving the opening and conclusion. This enables generating diverse dialogues for the same station. Dialogue generation proceeds in batches of up to four criteria, with up to three dialogue turns per batch, empirically chosen to balance generation quality and dialogue duration constraints. As OSCE stations are time-constrained, dialogue duration was estimated based on average speech rate and pauses.

\textbf{Physician generator:} The physician generator produces the physician's questions and statements. To ensure that these utterances remain coherent, contextually accurate, and aligned with the intended OSCE workflow, the prompt is dynamically enriched for each batch with the context of the physician sheet, the dialogue history, the current clinical stage, the relevant criteria checklist, and instructions for stage transitions when the batch manager signals a phase change.

\textbf{Patient generator:} 
Upon receiving a physician utterance, the retrieval module selects relevant information from the patient sheet as input context for the generator. If retrieval is disabled, the full sheet is provided. The generator then produces a context-aware response based on the patient sheet, instructions, and dialogue history.

\textbf{Controller module:} Generated responses are evaluated for fidelity to the patient sheet using an LLM-based scoring module that assesses consistency with the provided patient information and identifies potential contradictions. The controller assigns a score from 0 to 10 based on the number and severity of detected inconsistencies, along with a list of errors. Responses scoring above 8/10 are accepted, while lower-scoring responses are sent to a correction module before final validation. 

The \textbf{correction submodule} takes the flagged response along with an instruction prompt, the patient sheet, the physician’s question, and the errors identified by the controller. Its role is to correct the response by addressing the identified errors. Once corrected, the response is sent back to the controller module for final validation, forming a \textbf{reflection loop} that iteratively improves response fidelity and reduces contradictions with the patient sheet.

All modules are LLM-powered, and different models can be used per module. Examples are included in the prompt using a few-shot prompting strategy for both the controller and correction modules. Detailed prompts are provided in Appendix~\ref{appendix:prompts}.

\begin{table*}[htbp]
\centering
\small
\setlength{\tabcolsep}{3pt}
\resizebox{\textwidth}{!}{%
\begin{tabular}{l cc cc ccc}
\toprule
& \multicolumn{2}{c}{\textbf{Patient simulation}} & \multicolumn{2}{c}{\textbf{Physician performance}} & \multicolumn{3}{c}{\textbf{Linguistic quality}} \\
\cmidrule(lr){2-3} \cmidrule(lr){4-5} \cmidrule(lr){6-8}
\textbf{Model}
& \makecell{\textbf{Information} \\ \textbf{recall} \\ \scriptsize{(\%)}}
& \makecell{\textbf{Response} \\ \textbf{relevance} \\ \scriptsize{(/5)}}
& \makecell{\textbf{OSCE} \\ \textbf{evaluation} \\ \scriptsize{(\%)}}
& \makecell{\textbf{Role} \\ \textbf{adherence} \\ \scriptsize{(/5)}}
& \makecell{\textbf{Naturalness} \\ \scriptsize{(/5)}}
& \makecell{\textbf{Fluency} \\ \scriptsize{(/5)}}
& \makecell{\textbf{Coherence} \\ \scriptsize{(/5)}} \\
\midrule
Recorded dialogues & 43.58 & \underline{4.14} & 68.57 & 4.03 & 3.21 & 3.79 & 3.91 \\
\midrule
Claude-Haiku-4.5 & \textbf{52.24} & \underline{4.97} & \textbf{87.48} & \underline{4.76} & \underline{3.42} & \textbf{4.35} & \underline{4.28} \\
Gemini-3.1-Flash-Lite & 47.26 & 4.03 & \underline{85.96} & 4.67 & 3.40 & 4.07 & 4.20 \\
Ministral-14B & 47.44 & \underline{4.97} & 82.55 & 4.70 & 2.83 & 3.98 & 4.24 \\
\midrule
GPT-4o-mini & \underline{47.76} & \textbf{5.00} & 74.81 & \textbf{4.83} & \textbf{3.60} & \underline{4.31} & \textbf{4.42} \\
\bottomrule
\end{tabular}}
\caption{Results for recorded and generated dialogues on baseline configuration. Scores are computed using GPT-4o-mini as LLM-as-a-Judge and averaged over 44 generated dialogues. Best results in bold, second-best underlined.}
\label{tab:results_all}
\end{table*}

\section{Experiments and evaluation}
\subsection{Experimental protocol for dialogue generation}

To assess the contribution of each module, we define different experimental configurations: a baseline configuration without optional modules; the reflection loop only; and the reflection loop combined with retrieval. This allows us to evaluate the impact of each module on dialogue quality and controllability. When active, the reflection loop was configured with a maximum of 3 iterations and a minimum controller threshold score of 8/10 to accept a patient response (prompts available in Appendix~\ref{appendix:prompts}).

Configurations were evaluated using the following LLMs: Claude Haiku 4.5 \cite{anthropic2025haiku45}, Gemini-3.1-Flash-Lite \cite{google_gemini31flashlite_2026}, Ministral-14B-2512 \cite{ministral32026}, and GPT-4o-mini \cite{openai2024gpt4technicalreport} accessed through the OpenRouter API with a temperature of 0 and top-p of 0.9 across all modules (generation, retrieval, controller, corrector). Models were selected based on cost-efficiency considerations, as the system is intended for free use by medical students. Ministral-14B was additionally chosen for its potential for local deployment, an important factor for educational applications. 
We did not include domain-specialized medical LLMs as recent work suggests that specialized models do not necessarily outperform general-purpose ones on medical tasks \cite{dorfner2024biomedicallargelanguagesmodels, labrak_DrBenchmarkLargeLanguage_2024a,huang2026osce}. This is also consistent with our setting, where the VP is intended to simulate non-expert behavior. 

% CONFIGURATIONS : 1. baseline 2. reflection 3. retr. + refl. 4. 50% (x3) 5. 25% (x3)
% MODELS : 3
% STATIONS : 11
% DIALOGUES PER STATIONS : 4
% STATIONS * DIALOGUES * MODELS * CONFIG = 11*4*3*3 (1,2,3)*2 (4,5) = 792 dialogues
For each configuration and model, we generated 4 dialogues per station, yielding a total of 792 generated dialogues across 11 OSCE stations described in Appendix~\ref{appendix:case_annotations}.

\begin{table*}[htbp]
\centering
\small
\setlength{\tabcolsep}{3pt}
\resizebox{\textwidth}{!}{%
%\begin{tabular}{ll cc cc ccc}
\begin{tabular}{>{\centering\arraybackslash}m{2cm} l cc cc ccc}
\toprule
 & & \multicolumn{2}{c}{\textbf{Patient simulation}} & \multicolumn{2}{c}{\textbf{Physician performance}} & \multicolumn{3}{c}{\textbf{Linguistic quality}} \\
\cmidrule(lr){3-4} \cmidrule(lr){5-6} \cmidrule(lr){7-9}
 \makecell{\textbf{Physician}\\\textbf{performance}\\\textbf{variability}} & \makecell{\textbf{Experimental}\\\textbf{setup}}
& \makecell{\textbf{Information} \\ \textbf{recall} \\ \scriptsize{(\%)}}
& \makecell{\textbf{Response} \\ \textbf{relevance} \\ \scriptsize{(/5)}}
& \makecell{\textbf{OSCE} \\ \textbf{evaluation} \\ \scriptsize{(\%)}}
& \makecell{\textbf{Role} \\ \textbf{adherence} \\ \scriptsize{(/5)}}
& \makecell{\textbf{Naturalness} \\ \scriptsize{(/5)}}
& \makecell{\textbf{Fluency} \\ \scriptsize{(/5)}}
& \makecell{\textbf{Coherence} \\ \scriptsize{(/5)}} \\
\midrule
\multirow{3}{*}{{100\%}}
& baseline     & 52.24 & \textbf{4.97} & 87.48 & 4.76 & 3.42 & \textbf{4.35} & 4.28 \\
& reflection & 54.29 & 4.83 & 87.75 & 4.85 & 3.57 & 4.30 & \textbf{4.36} \\
& refl. + retr.  & \textbf{56.62} & 4.89 & \textbf{89.21} & \textbf{4.89} & \textbf{3.59} & 4.29 & \textbf{4.36} \\
\midrule
\multirow{3}{*}{{50\%}}
& baseline     & 24.12 & 4.85 & \textbf{46.21} & 4.31 & \textbf{3.52} & 4.22 & 4.26 \\
& reflection & \textbf{30.84} & \textbf{5.00} & 45.59 & \textbf{4.55} & 3.51 & \textbf{4.24} & \textbf{4.27} \\
& refl. + retr.  & 27.23 & 4.86 & 45.05 & 4.14 & 3.46 & 4.20 & 4.20 \\
\midrule
\multirow{3}{*}{{25\%}}
& baseline     & 17.80 & 4.44 & 25.69 & 3.89 & 3.00 & 4.00 & \textbf{4.8} \\
& reflection & \textbf{22.65} & 4.67 & \textbf{29.89} & \textbf{4.17} & 3.35 & \textbf{4.13} & 4.15 \\
& refl. + retr.  & 18.74 & \textbf{4.75} & 27.06 & 4.00 & 3.00 & 4.00 & 4.10 \\
\bottomrule
\end{tabular}
}
\caption{Results for Claude-Haiku-4.5 across three experimental setups (baseline, reflection loop, and reflection + retrieval) under varying physician performance constraints (100\%, 50\%, and 25\% OSCE criteria coverage). Best results for each physician performance variation in bold.} %Scores are computed using GPT-4o-mini as LLM-as-a-Judge and averaged over 44 generated dialogues.}
\label{tab:results_50students}
\end{table*}

\subsection{Evaluation setup: description and metrics}

All evaluations were conducted in an LLM-as-a-Judge framework \cite{gu_SurveyLLMasajudge_2026}, using GPT-4o-mini \cite{openai2024gpt4technicalreport} with a temperature of 0. We assess generated and recorded dialogues along three dimensions, each comprising multiple metrics. Evaluation prompts are available in Appendix~\ref{appendix:prompts_eval}. To assess the reliability of this automatic evaluation, we additionally collected a small set of annotations from 6 annotators without medical training.

\paragraph{1. Patient simulation quality} evaluates how faithfully and appropriately the VP behaves with respect to the patient sheet. We propose the following metrics for the patient simulation quality evaluation:
 
\noindent\hspace{0.5em}\textbf{$\bullet$ Information recall} measures the proportion of patient sheet elements mentioned by the VP during the dialogue. In a preliminary step, we use GPT-4o-mini to transform each patient sheet into a binary checklist (e.g., mentions the patient's name, reports allergies). The LLM-based judge then reviews all patient utterances and marks each item as mentioned or not, yielding a percentage score.

\noindent\hspace{0.5em}\textbf{$\bullet$ Response relevance} assesses whether patient utterances are appropriate given the physician's questions and the patient sheet content. The LLM-based judge assigns a score from 1 (very poor) to 5 (excellent).

\paragraph{2. Physician performance} \label{sec:student_perf} evaluates the quality of the physician's conduct during the consultation, with the following metrics:

\noindent\hspace{0.5em}\textbf{$\bullet$ OSCE evaluation} implements traditional OSCE scoring: the evaluator sheet lists specific criteria that the physician is expected to address (e.g., asks about allergies, proposes a treatment), and the judge checks them as met or not, yielding a percentage score.

\noindent\hspace{0.5em}\textbf{$\bullet$ Role adherence} assesses if the physician appropriately fulfilled the role defined in the physician sheet: respecting the clinical context, staying within the OSCE framework, and behaving professionally. Only physician utterances are evaluated. The LLM-based judge assigns a score from 1 (very poor, major deviations from the expected role) to 5 (excellent, perfect role adherence).

\paragraph{3. Linguistic quality} evaluates the dialogue independently of its clinical content, focusing on three standard dialogue quality metrics. \textbf{Naturalness} measures whether the dialogue resembles a real spoken conversation between two people. \textbf{Fluency} assesses the smoothness and grammatical correctness of the exchanges. \textbf{Coherence} evaluates the logical structure and progression of the dialogue. Each criterion is scored on a 0–100 scale using detailed rubrics we defined in prompts, then rescaled to a 1–5 Likert scale for consistency with the other similar metrics in this framework (e.g., response relevance). All evaluation prompts are available in Appendix~\ref{appendix:prompts_eval}.

\section{Results and analysis}
We conduct four analyses: (1) comparison between recorded and baseline-generated dialogues across the four models (\autoref{tab:results_all}), (2) impact of reflection and retrieval modules with Claude-Haiku-4.5 (\autoref{tab:results_50students}), (3) robustness to varying physician performance (\autoref{tab:results_50students}), and (4) station-level naturalness analysis (\autoref{fig:naturalness}). Detailed results are provided in Appendix~\ref{appendix:detailed_results} with all configurations and models.

\subsection{Generated vs. recorded dialogues}
Across all evaluation dimensions, generated dialogues consistently outperform recorded interactions, including higher information recall across all models, as shown in \autoref{tab:results_all}. Linguistic quality scores are also higher for generated dialogues, except for Ministral-14B, whose naturalness score falls below the recorded dialogue baseline of 3.21, suggesting that this model produces less conversationally realistic utterances. 

This gap is partly expected: recorded dialogues reflect real student variability, including incomplete coverage of OSCE criteria, hesitations, and communication difficulties, whereas generated dialogues benefit from systematic criterion coverage and controlled patient behavior. Additionally, these results may partly reflect biases in LLM-as-a-Judge evaluation, which could favor the style and structure of LLM-generated text over the disfluencies and variability characteristic of spontaneous speech.  

As part of the linguistic quality evaluation (Section~\ref{sec:agg}), annotators were asked to determine whether each dialogue was LLM-generated or recorded from a real interaction. They correctly identified the dialogue origin in only 14 of 24 judgments (58.3\%), achieving identical accuracy for LLM-generated and recorded dialogues. %Annotators frequently associated repeated phrases, role reversals, and overly formal language with AI-generated dialogues, while disfluencies and topic drift were perceived as signs of human interactions.

For completeness, we also include the results of GPT-4o-mini, despite using it as the LLM-as-a-Judge evaluator. Interestingly, it does not achieve the highest scores across the evaluated models, despite also serving as the evaluator. This observation, however, should not be interpreted as evidence that the evaluation is free from model-specific biases.

Since Claude-Haiku-4.5 consistently achieves the best overall performance across evaluation dimensions, we use this model for the remaining experiments. 

\subsection{Impact of the reflection and retrieval modules}

Table~\ref{tab:results_50students} shows the contribution of reflection loop and retrieval module compared to the baseline system with Claude-Haiku-4.5. Overall, the reflection loop consistently improves patient simulation metrics compared to the baseline, particularly information recall, while maintaining stable linguistic quality. These improvements suggest that the controller and corrector modules help refine responses and reduce inconsistencies in generated dialogues. The reflection loop was triggered in 3.6\% of patient turns, indicating that most responses were accepted by the controller without requiring correction. When activated, the correction process improved the controller score in 79.5\% of cases.

Adding retrieval sometimes further improves performance, yielding the highest information recall and OSCE evaluation scores. However, these gains remain inconsistent across setups, suggesting that retrieval effectiveness depends on the interaction dynamics. Across configurations, linguistic quality remains stable, indicating that both modules improve fidelity and task performance without degrading dialogue naturalness. Overall, the reflection loop and retrieval modules sometimes improve results, but their impact remains moderate. 

\begin{figure}[t]
\centering
\includegraphics[width=1\linewidth]{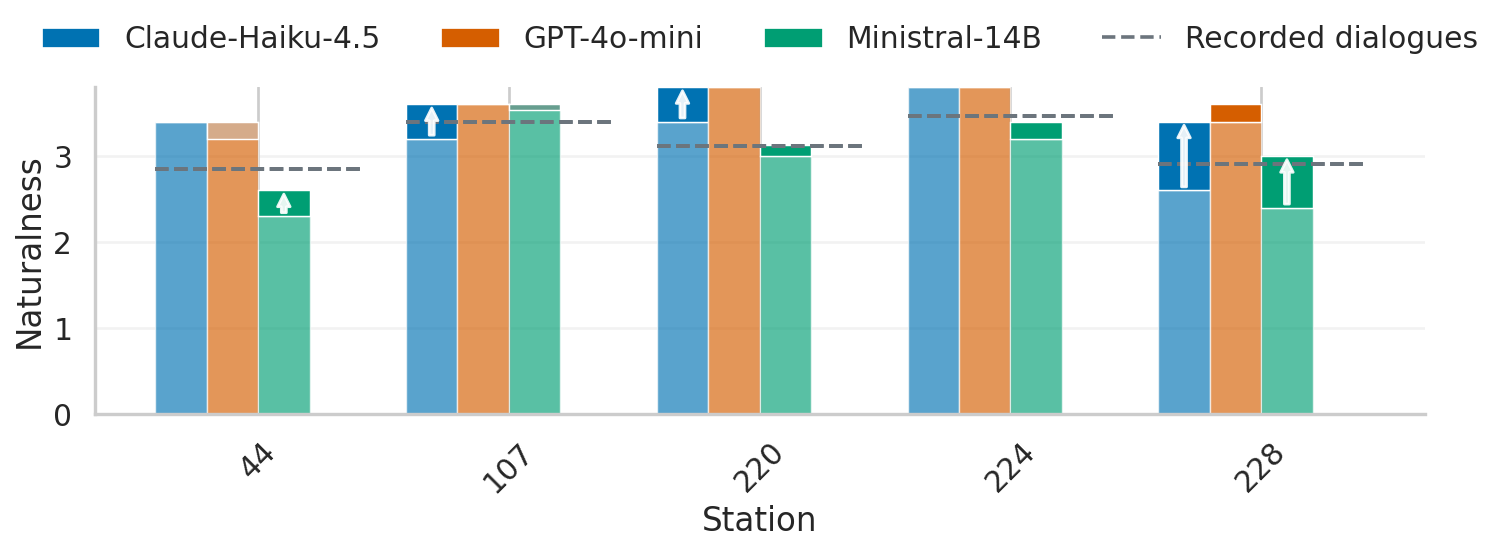}
\caption{Naturalness score across all models and recorded dialogues: bars represent baseline performance averaged over four generated dialogues, deltas indicate performance changes with the retrieval and reflection loop for five representative stations.}
\label{fig:naturalness}
\end{figure}

To further assess whether these improvements affect conversational quality at the utterance level, we evaluate linguistic quality scores separately for each speaker role (\autoref{tab:linguistic_quality_full}). Linguistic quality remains stable across experimental setups for both speaker roles. Physician scores are consistently slightly higher than patient simulation scores, suggesting that patient generation may introduce more variability than physician utterances.

\begin{table*}[ht]
\centering
\small
\resizebox{\textwidth}{!}{
\begin{tabular}{lcccccc}
\toprule
& \multicolumn{3}{c}{\textbf{Patient simulation}}
& \multicolumn{3}{c}{\textbf{Physician performance}} \\
\cmidrule(lr){2-4} \cmidrule(lr){5-7}
\makecell[l]{\textbf{Experimental}\\\textbf{setup}}
& \makecell{Naturalness\\\scriptsize(/5)}
& \makecell{Fluency\\\scriptsize(/5)}
& \makecell{Coherence\\\scriptsize(/5)}
& \makecell{Naturalness\\\scriptsize(/5)}
& \makecell{Fluency\\\scriptsize(/5)}
& \makecell{Coherence\\\scriptsize(/5)} \\
\midrule
baseline       & \textbf{3.82} & \textbf{4.26} & \textbf{4.25} & \textbf{3.80} & \textbf{4.40} & \textbf{4.65} \\
reflection     & 3.78          & 4.23          & 4.10          & 3.67          & 4.31          & 4.47          \\
refl.\ + retr. & \textbf{3.82} & 4.25          & \textbf{4.25} & 3.68          & 4.34          & 4.48          \\
\bottomrule
\end{tabular}
}
\caption{Linguistic quality scores per speaker role for Claude-Haiku-4.5, evaluated with GPT-4o-mini as LLM-as-a-Judge. Full dialogues are provided to the judge, which scores each speaker independently. Scores are computed on a 0–100 scale using rubric-based prompts and rescaled to 1–5 Likert scores.}
\label{tab:linguistic_quality_full}
\end{table*}

\subsection{Robustness to physician performance variability}
 
To assess our VP robustness to variable physician performance, we degraded the physician generator by providing only 50\% or 25\% of the expected OSCE evaluation criteria, and perturbing them following \citet{huang2026osce}, reported in~\autoref{tab:results_50students}. This degradation led to a drop in information recall score, which is consistent with the statistically significant correlation between information recall and OSCE scores, reported in Appendix~\ref{appendix:correlation_recallosce}: as the physician asks fewer questions, the patient provides less information.

Response relevance remains stable and linguistic quality scores slightly decrease. Compared to the 100\%-criteria setting, these results suggest that the VP remains robust to variations in physician performance, particularly when using the retrieval and reflection loop pipeline, which maintains high response relevance with degraded interaction conditions. This robustness allows the VP to adapt to student performance and enables learning even from failed interviews.

\subsection{Station-level analysis}
\autoref{fig:naturalness} presents the analysis of naturalness scores at the station level to assess where the reflection loop and retrieval module are most beneficial. We focus on five representative stations (44, 107, 220, 224, 228) covering diverse clinical contexts (see Appendix~\ref{appendix:case_annotations}), including different specialties and objectives. The gain on station 228 may be partly driven by few-shot prompting, as one example directly matches this station. Overall, results suggest that improvements in naturalness are very context-dependent and not uniform across stations. This variability further suggests that a discretized station design, based on more static and well-defined categories, could improve simulation consistency.

\subsection{Agreement between human evaluators and LLM-as-a-Judge} \label{sec:agg}

To assess the reliability of our LLM-as-a-Judge framework, six annotators each evaluated four dialogues on three linguistic-quality criteria (naturalness, fluency, and coherence) using the same 1--5 Likert scale as the LLM. Each of the 12 dialogues was rated by two annotators.

\begin{table}[H]
\centering
\small
\setlength{\tabcolsep}{4pt}
\begin{tabular}{l ccc}
\toprule
\textbf{Criterion} & \textbf{ICC (human)} & \textbf{Pearson} & \textbf{Spearman} \\
\midrule
Naturalness & $0.131$  & $0.08$        & $0.09$  \\
Fluency     & $-0.134$ & $0.46$        & $0.45$  \\
Coherence   & $-0.040$ & $0.76^{\ast}$ & $0.45$  \\
\bottomrule
\end{tabular}
\caption{Inter-annotator agreement and correlation between mean human scores and LLM-as-a-Judge scores ($n=12$ dialogues). $^{\ast}p<0.05$}
\label{tab:icc-correlation}
\end{table}

Intraclass Correlation Coefficient (ICC) \cite{koo2016guideline} is weak across all three criteria, reflecting the subjective nature of linguistic-quality evaluation. Human-LLM agreement remains limited, with only coherence showing a significant Pearson correlation, as shown in~\autoref{tab:icc-correlation}.

We also evaluated agreement between GPT-4o-mini and human evaluator scores on the OSCE evaluation metric using ICC and Spearman correlation. On 67 recorded dialogues across 14 OSCE stations, agreement was moderate (ICC = 0.41, $\rho = 0.33$), and became strong when restricting the analysis to patient-only stations without document analysis (ICC = 0.85, $\rho = 0.82$). These results suggest that the LLM-as-a-Judge is more reliable for checklist-style OSCE evaluation than for linguistic-quality assessment, although these results remain preliminary.
%Other station types yielded inconclusive results because of limited sample sizes ($n \leq 4$). 
%These results do not allow us to conclude that the LLM-as-a-Judge is reliable: agreement appears stronger on checklist-style OSCE criteria with larger samples, but the linguistic-quality results rest on a small annotated sample and should be read as preliminary rather than as evidence of reliability.

\section{Conclusion}
We introduced a French OSCE dialogue dataset combining 240 student-patient OSCE training dialogues and 792 generated dialogues, providing a new data resource for French medical education. We proposed a controllable LLM-based generation pipeline with retrieval and controller modules to improve patient fidelity and dialogue coherence, and an LLM-as-a-Judge evaluation framework assessing patient simulation quality, physician performance, and linguistic quality.

Results show that generated dialogues consistently outperform recorded interactions in patient simulation quality, particularly in terms of information recall and response relevance. The proposed VP system with a reflection loop yields small gains in fidelity-related metrics, while retrieval provides additional but less consistent improvements depending on interaction context and model. Our VP remains stable across variations in physician performance, while station-level analyses reveal scenario-dependence in naturalness, suggesting the need for more fine-grained station-level modeling to better adapt system behavior.

Future work will include extending the pipeline to document-based stations, improving controllability of VP behavior, and validating our proposed VP system in real training conditions with medical students.

\newpage

\section*{Limitations \& Ethics statement}
\paragraph{Limitations} This study has several limitations. First, while our dataset includes 192 OSCE stations, only 23 were recorded, and dialogue generation was conducted on a subset of 11 stations. This limits the diversity of clinical scenarios covered and the generalization of our findings, though the dataset is under active expansion. Second, our evaluation relies entirely on a single LLM-as-a-Judge with a single evaluation run per dialogue. Although we validated the OSCE evaluation criterion against human annotations, the remaining criteria lack human validation, and inter-run variability was not assessed. Third, the generation pipeline was not tested in real training conditions with medical students interacting with the VP, leaving its pedagogical effectiveness still unevaluated. Fourth, the pipeline relies heavily on LLM calls across all modules (generation, retrieval, controller, corrector, evaluation), which raises concerns regarding reproducibility, cost, and scalability. We estimate the generation cost of a single dialogue at approximately \$0.20, but a full experimental run involves numerous API calls across multiple modules. Finally, only three models were evaluated; larger or domain-specialized medical models may yield different results. Future work will focus on expanding human evaluation, assessing inter-annotator agreement, conducting user studies with medical students, and evaluating the pipeline in real training conditions.

\section*{Ethics statement}
This study follows ethical guidelines for the use of language technologies in educational and clinical training contexts. No personal, clinical, or patient-identifiable data were collected or processed. All real interactions used in this work were recorded with appropriate authorization, and all scenarios were fictitious and designed in a game-like educational setting, ensuring full anonymization of participants. We emphasize that the goal of this work is to support the development of assistive technologies that augment human learning without reducing expertise or learner agency, particularly in sensitive domains such as medical education.
%We acknowledge that our approach is computationally intensive due to extensive use of LLMs, leading to non-negligible computational and environmental costs. However, we consider this trade-off acceptable given the controlled experimental framework it enables and the fact that we could estimate +3.6 surcoût en itération. Improving efficiency through model reuse and optimization remains an important direction for future work.

\section{Acknowledgements}
We thank the members of the Multispeech team for their contributions to the manual transcription and human evaluation of the dialogues used in this study. We also acknowledge ECNAsso for facilitating data collection. This work benefited from government funding managed by the National Research Agency under France 2030 via the ENACT AI Cluster (ANR-23-IACL-0004), and from support by Région Grand Est.

\bibliography{tom,custom,references_zotero_vpatient}

\clearpage
\appendix
\section{Appendix}
\label{sec:appendix}

\subsection{Data collection: recording protocol} \label{appendix:recording}

\paragraph{Recording context:} Students followed a predefined schedule for the training sessions, rotating through numbered stations. Students playing patient roles received detailed station descriptions in advance and often portrayed the same patient throughout a session, appearing in multiple recordings for the same station, while students playing physicians accessed station-specific context only upon entering the room. An evaluator, also a medical student, observed the interaction and provided structured feedback using the evaluator sheet.

\paragraph{Recording protocol:} The recordings were conducted unobtrusively during the training sessions. Out of the eight available examination rooms, two were equipped for audio recording. In each recorded rooms, wireless clip-on microphones (DJI Mic Mini), worn by both the evaluated student and the simulated patient, were used to record. In addition to audio, 85 of the recorded dialogues include video data, which were not used in the present work. The collected data were transferred and stored on encrypted hard drives with restricted access. 

\subsection{Transcription evaluation: WER computation} \label{appendix:wer}

The WER is defined as:
\[
\mathrm{WER} = \frac{S + D + I}{N},
\]
where $S$ denotes the number of substitutions, $D$ the number of deletions, $I$ the number of insertions, and $N$ the total number of words in the reference transcription.

Before WER computation, transcripts were normalized and tokenized using the \textit{spaCy} pipeline with the \texttt{fr\_core\_news\_md} model \cite{spacy_fr_core_news_md}. WER computation was performed using a custom Python pipeline based on the \texttt{jiwer} library \cite{jiwer}. Automatic and reference transcripts were paired by dialogue identifier, normalized, segmented by speaker, and then evaluated.

\subsection{Correlation test: information recall and OSCE score}
\label{appendix:correlation_recallosce} 
Pearson (r = 0.29) and Spearman (r = 0.31) correlations between information recall and OSCE score were computed, showing a weak but statistically significant positive association (p $<$ 0.001). This indicates a partial relationship between information recall and OSCE performance, while suggesting that other conversational and clinical reasoning factors also contribute to the overall score. Information recall is reported for informational purposes only, as exhaustive disclosure of patient information is neither expected nor desirable in OSCE settings.

\newpage
\onecolumn
\subsection{Dataset details} \label{appendix:case_annotations}
\begin{table*}[h!]
\centering
\small
\begin{tabularx}{\textwidth}{c l l X}
\hline
\textbf{Station (ID)} & \textbf{Specialty} & \textbf{Consultation type} & \textbf{Objective} \\
\hline
44  & Psychiatry          & Emergency             & History taking; diagnosis; clinical summary \\
46  & Pediatrics          & First consultation     & Breaking bad news; patient education \\
107 & General practice    & Follow-up              & History taking; diagnosis \\
120 & Neurology           & Follow-up              & Breaking bad news; patient education \\
220 & Cardiology; ICU     & Emergency              & Diagnosis; interprofessional coordination \\
224 & Pulmonology   & Follow-up              & History taking; diagnosis; clinical summary \\
225 & Occupational medicine & Follow-up            & Specialist referral; patient education \\
226 & General practice    & Follow-up              & Patient education; treatment initiation \\
228 & General practice     & Emergency              & History taking; diagnosis; specialist referral \\
233 & General practice     & Follow-up              & History taking; prescription renewal; patient education \\
235 & Emergency practice & Emergency              & Examination interpretation; history taking; diagnosis; patient education \\
\hline
\end{tabularx}
\caption{Annotations of the 11 OSCE stations used for dialogue generation, including medical specialty, consultation type, and communication objectives.}
\end{table*}
\FloatBarrier

\subsection{Detailed results} \label{appendix:detailed_results}
\begin{table*}[htpb]
\centering
\small
\setlength{\tabcolsep}{3pt}
\resizebox{\textwidth}{!}{%
\begin{tabular}{ll cc cc ccc}
\toprule
& & \multicolumn{2}{c}{\textbf{Patient simulation}} & \multicolumn{2}{c}{\textbf{Physician performance}} & \multicolumn{3}{c}{\textbf{Linguistic quality}} \\
\cmidrule(lr){3-4} \cmidrule(lr){5-6} \cmidrule(lr){7-9}
\textbf{Model} & \textbf{Setup}
& \makecell{\textbf{Information} \\ \textbf{recall} \\ \scriptsize{(\%)}}
& \makecell{\textbf{Response} \\ \textbf{relevance} \\ \scriptsize{(/5)}}
& \makecell{\textbf{OSCE} \\ \textbf{evaluation} \\ \scriptsize{(\%)}}
& \makecell{\textbf{Role} \\ \textbf{adherence} \\ \scriptsize{(/5)}}
& \makecell{\textbf{Naturalness} \\ \scriptsize{(/5)}}
& \makecell{\textbf{Fluency} \\ \scriptsize{(/5)}}
& \makecell{\textbf{Coherence} \\ \scriptsize{(/5)}} \\
\midrule
Recorded dialogues & & 43.58 & 4.14 & 68.57 & 4.03 & 3.21 & 3.79 & 3.91 \\
\midrule
\multirow{3}{*}{Claude-Haiku-4.5}
& baseline     & 52.24 & \textbf{4.97} & 87.48 & 4.76 & 3.42 & \textbf{4.35} & 4.28 \\
& reflection & 54.29 & 4.83 & 87.75 & 4.85 & 3.57 & 4.30 & \textbf{4.36} \\
& refl. + retr.  & \textbf{56.62} & 4.89 & \textbf{89.21} & \textbf{4.89} & \textbf{3.59} & 4.29 & \textbf{4.36} \\
\midrule
\multirow{3}{*}{GPT-4o-mini}
& baseline     & \textbf{47.76} & \textbf{5.00} & 74.81 & \textbf{4.83} & 3.60 & \textbf{4.31} & 4.42 \\
& reflection & 44.66 & 4.91 & \textbf{76.79} & 4.82 & \textbf{3.64} & 4.29 & \textbf{4.44} \\
& refl. + retr.  & 42.39 & 4.97 & 75.71 & 4.69 & 3.59 & 4.30 & 4.38 \\
\midrule
\multirow{3}{*}{Ministral-14B}
& baseline     & 47.44 & 4.97 & 82.55 & 4.70 & 2.83 & 3.98 & 4.24 \\
& reflection & 50.02 & \textbf{4.98} & \textbf{84.44} & \textbf{4.74} & \textbf{2.95} & \textbf{4.03} & \textbf{4.28} \\
& refl. + retr.  & \textbf{52.40} & 4.95 & 83.50 & 4.55 & 2.60 & 3.96 & 4.21 \\
\bottomrule
\end{tabular}
}
\caption{Evaluation results across all three dimensions with GPT-4o-mini as LLM-as-a-Judge, for all three models and each configuration: baseline, reflection loop only, reflection loop + retrieval. Each score is averaged over 44 generated dialogues. Best results for each model in bold.} %Baseline = no optional modules; + reflection = with reflection loop; + retrieval = with reflection and retrieval module.}
\end{table*}
\FloatBarrier

\newpage
\twocolumn
\subsection{Prompts for dialogue generation} \label{appendix:prompts}
% \onecolumn
\begin{lstlisting}[title=\textbf{OSCE criteria classification prompt}]
You are a medical education expert. Classify each of the following clinical consultation criteria into exactly one of these stages:

{stages_desc}

Criteria to classify:
{criteria_text}

Return ONLY a JSON object mapping criterion number to stage id:
{{
  "1": "opening_and_preparation",
  "2": "information_collection",
  ...
}}

No explanation, no markdown fences, just the JSON object.
\end{lstlisting}

\begin{lstlisting}[title=\textbf{Physician's utterances generation prompt}]
You are generating the responses of a physician during an Objective Structured Clinical Examination. The exam takes place in a French medical university, and entirely in French. You must interact with the patient, asking questions and making comments as a real physician would do during a consultation. You must strictly follow the instructions and context provided below.

CONTEXT AND INSTRUCTIONS:
{consultation_context}

CURRENT PHASE: {stage_label} (batch {batch_number}/{total_batches})
{stage_transition_instruction}

OBJECTIVES FOR THIS PHASE:
{current_checklist}

{covered_criteria_section}

CONVERSATION HISTORY:
{conversation_history}

LAST PATIENT RESPONSE:
{last_patient_response}

You have 8 minutes total. Currently, {time_info} remains. If less than 2 minutes remain, prioritize closing the consultation and addressing the most critical objectives.
{final_part_instruction}

GENERAL RULES:
- Output speech only (no parentheses, no actions, no thoughts).
- Contextual fidelity: Strictly follow the role, objectives, and constraints provided above.
- Use open-ended questions to explore symptoms. 
- Medical history: Investigate relevant medical history when appropriate.
- Adaptability: Tailor your follow-up questions to the patient's specific answers. Do not produce long answers. Be concise.
- Flow control: Ask only one question at a time.
- Coverage: Make sure to address ALL criteria listed in OBJECTIVES FOR THIS PHASE before the conversation moves on.
- Phase awareness: Your questions should be appropriate for the current phase of the consultation.

IMPORTANT: Only generate the physician's next spoken response as text. Do not include stage directions, actions, or any narration. NO PHYSICAL EXAMINATIONS. Your answer must be short (1-2 sentences).

Physician's next question:
\end{lstlisting}

\begin{lstlisting}[title=\textbf{Patient's utterrances generation prompt}]
You are generating the responses of a patient during an Objective Structured Clinical Examination, to help teach medicine students. The exam takes place in a French medical university, and entirely in French.
You must interact with and answer the doctor's questions as truthfully as possible.
Compose your responses as though you are speaking with the doctor-student. Try to be as concise as possible, as the exam only lasts 7 minutes. Keep answers short and natural. Use everyday language, not medical jargon.

Rules:
- Answer only what is asked, and give one small piece of information at a time.
- Do not volunteer extra details; wait for the doctor to ask.
- If you don't know or the information is not in the patient data, say you don't know.
- If the doctor uses complex terms, ask for a simpler explanation.
- Do not argue with the doctor; if you are unsure, ask for clarification.
- Keep responses to 1 short sentence or two short sentences at most.
- Do not repeat the same closing line over and over; once you agree or acknowledge, wait for the next question.
- Remember what the doctor just told you and respond consistently, even if it changes your attitude.
- Output speech only (no parentheses, no actions, no thoughts).
- Once the interview is coming to a close, and it is your last turn, thank the doctor and say goodbye.

{patient_data_section}

Start the dialogue with "phrase_demarrage" indicated in your patient profile.

Conversation history:
"{conversation_history}"

Last physician message: 
"{doctor_question}"

Your answer: 
\end{lstlisting}

\begin{lstlisting}[title=\textbf{Patient's utterance evaluation (controller) prompt}]
You are an expert evaluator in clinical communication for OSCE (Objective Structured Clinical Examination) exams.
Your task is to evaluate a simulated patient reply written in plain natural language.
The response you evaluate is NOT a JSON and MUST be treated as raw text only.

PATIENT PROFILE:
{patient_profile}

HISTORY:
{conversation_history}

GENERATED RESPONSE:
{patient_response}

EVALUATION CRITERIA:
- Consistency: The response is compatible with the patient profile, without contradictions or invented information.
- Patient realism: The level of language and understanding matches that of a patient.
- Credible interaction: Only verbal discourse; no actions or emotions described; concise response.

CRITICAL CONCISION:
- The patient answers accordingly to their profile and to the doctor's answer. 
- The response MUST NOT contain asterisks (*), parentheses to describe actions, or descriptions of gestures/facial expressions. ANY text between asterisks (action) or parentheses (gesture) is a MAJOR ERROR that makes the response NON-COMPLIANT ("conforme": false).
- Patient role: The patient does not reason like a doctor and does not suggest diagnoses or treatments.
- Linguistic quality: The response is clear, natural, and correct.
- If the answer is empty, give a maximum score of 1. 


STRICT RULE ON ASTERISKS:
If the response contains even a SINGLE asterisk (*) or parenthesis describing an action:
  "conforme": false
  "score": maximum 4
  Add to errors: "Presence of asterisks or prohibited action descriptions".

YOUR MISSION:
- Verify the absence of asterisks or parentheses describing actions.
- Detect any inconsistency or role error.
- Identify specific issues.

STRICT OUTPUT FORMAT (JSON):
  {{
  "conforme": true | false,
  "score": int from 0 to 10,
  "erreurs": [
  "precise description of problem 1",
  "precise description of problem 2"
  ]
  }}

RULES:
- Produce ONLY this JSON, nothing else.
- No text before or after the JSON.
- No markdown tags like ```json.
- No comments.
- ONLY ONCE (do not repeat the JSON).

Evaluation:
\end{lstlisting}

\begin{lstlisting}[title=\textbf{Patient's utterances correction prompt}]
You are a medical simulation assistant. Your task is to correct a simulated patient's response to ensure it is realistic, coherent with the patient profile, and follows the format rules.

PATIENT PROFILE:
{patient_profile}

DOCTOR'S QUESTION:
{doctor_question}

CURRENT PATIENT RESPONSE (TO CORRECT):
{patient_response}

ERRORS DETECTED:
{errors_text}

--------------------------------------------------
CORRECTION INSTRUCTIONS
--------------------------------------------------

1. **MINIMAL CHANGES**: Make ONLY the changes necessary to fix the listed errors
   - Keep the same tone, style, and structure
   - Keep the same vocabulary level
   - Change ONLY what is incorrect

2. **FORBIDDEN ELEMENTS** (must be removed):
   - Asterisks for actions: *soupir*, *grimace*, *acquiesce*
   - Parentheses for actions: (sourit), (hesite), (reflechit)
   - Physical descriptions: "Je hoche la tete", "Je croise les bras"
   - Stage directions or narration

3. **ALLOWED ELEMENTS**:
   - Verbal hesitations: "euh...", "hum...", "ben...", "alors..."
   - Natural speech patterns: repetitions, corrections, incomplete sentences
   - Simple words and everyday language (patients are not doctors)

4. **COHERENCE WITH PROFILE**:
   - If the profile says the patient takes Metformin, the response must mention it
   - If the profile says non-smoker, the response must NOT mention smoking
   - Respect all factual elements from the profile

5. **NATURAL SPEECH**:
   - Keep responses SHORT (1-2 sentences typical for a single turn)
   - Patients don't know exact medical terms or dosages


YOUR TASK:
Provide ONLY the corrected patient response. No explanations, no comments, no markdown, just the corrected text.


CORRECTED RESPONSE:
\end{lstlisting}

\subsection{Prompts for dialogue evaluation} \label{appendix:prompts_eval}
% \vspace{0.5em}
\begin{lstlisting}[title=\textbf{Patient simulation quality evaluation prompt: information recall}]
You are an expert medical evaluator. For each criterion, determine whether the patient mentioned it in the dialogue.
Respond ONLY with valid JSON.

Patient turns only:
---
{patient_turns}
---

Binary criteria:
{criteres_json}

Return:
{{
  "evaluation_binaire": [
    {{"id": <id>, "critere": "<text>", "mentionne": true/false, "justification": "<short extract or reason>"}}
  ],
  "score_binaire": {{"mentionne": <n>, "total": <n>, "pourcentage": <float>}}
}}

RULES:
- Return ONLY valid JSON.
- Evaluate ONLY patient speech.
- Do NOT penalize date inconsistencies.
\end{lstlisting}

\begin{lstlisting}[title=\textbf{Patient simulation quality evaluation prompt: response relevance}]
You are a medical dialogue evaluation expert.
Evaluate whether patient responses are relevant to the doctor's questions and consistent with the patient sheet.
Respond ONLY with valid JSON.

Patient sheet:
{fiche_json}

Patient turns:
{patient_turns}

Dialogue for context:
{dialogue}

Evaluate whether patient responses are relevant to:
1. The doctor's questions
2. The patient sheet

Score relevance:
1 = Very poor  
2 = Poor  
3 = Acceptable  
4 = Good  
5 = Excellent  

Return:
{{
  "score_response_relevance": <1-5>,
  "justification": "<brief explanation>",
  "examples": {{
    "relevant": ["..."],
    "irrelevant": ["..."]
  }}
}}

Rules:
- Evaluate ONLY patient responses
- Compare responses with both doctor questions and patient sheet
- Do not judge medical correctness beyond the sheet
- Return ONLY valid JSON
\end{lstlisting}

%\begin{lstlisting}[title=\textbf{Patient simulation quality evaluation prompt: patient faithfulness}]
%\end{lstlisting}

\begin{lstlisting}[title=\textbf{Physician performance evaluation prompt: OSCE evaluation}]
You are an expert medical evaluator. For each criterion, determine whether the physician mentioned or addressed it in the dialogue.
Respond ONLY with valid JSON.

Full dialogue:
{dialogue}

Binary criteria to evaluate:
{fiche_doctor}

Return:
{{
  "osce_evaluation": [
    {{"id": <id>, "critere": "<text>", "ok": true/false, "justification": "<short extract or reason>"}}
  ],
  "score_osce": {{"ok": <n>, "total": <n>, "pourcentage": <float>}}
}}

RULES:
- Return ONLY valid JSON.
- Evaluate ONLY physician speech.
\end{lstlisting}

\begin{lstlisting}[title=\textbf{Physician performance evaluation prompt: role adherence}]
You are a medical simulation expert. Evaluate whether the physician correctly plays their role given the context provided to them.
Respond ONLY with valid JSON.

Context given to the doctor at the start of the simulation:
---
{medecin_context}
---

Full dialogue:
---
{dialogue}
---

Based solely on the context above, evaluate the physician's overall role adherence. Consider the following aspects in your assessment:
- Did the physician respect the clinical context they were given?
- Was the communication style appropriate for a remote medical consultation?
- Did the physician stay within the bounds of what is possible in a spoken interaction?

Score on a 1-5 Likert scale:
- 1 = Very poor: Major deviations from role, inappropriate actions, context ignored
- 2 = Poor: Several deviations from role or significant inappropriate behaviors
- 3 = Acceptable: Mostly in role with some notable issues
- 4 = Good: Consistently in role with only minor issues
- 5 = Excellent: Perfect role adherence, fully respects context and constraints

Return:
{{
  "role_adherence": {{
    "score": <1-5>,
    "justification": "<detailed explanation of the score>",
    "issues": ["<list any problems found>"]
  }},
  "score_role_adherence": <1-5>
}}

RULES:
- Return ONLY valid JSON.
- Evaluate ONLY physician speech.
- PENALIZE if the physician starts a physical exam or mimics actions impossible in a remote simulation (e.g. shaking hands, auscultation).
- PENALIZE if the physician describes emotions or physical gestures (e.g. *smiling*, *looking concerned*), as this breaks immersion.
- Judge the physician ONLY against the context they were given, not against hidden patient information.
\end{lstlisting}

\begin{lstlisting}[title=\textbf{Linguistic quality evaluation prompt: naturalness, fluency, coherence}]
You are an expert evaluator of medical dialogue linguistic quality. You evaluate dialogues on three dimensions: naturalness, fluency, and coherence.
You use strict rubrics and score on a 0-100 scale.
Respond ONLY with valid JSON.

Evaluate the following dialogue on naturalness, fluency, and coherence.

**Rubrics**:
{rubrics}

**Dialogue to evaluate**:
---
{dialogue}
---

Return:
{{
  "linguistic_quality": {{
    "naturalness": {{
      "score_100": <0-100>,
      "band": "<0-20|20-40|40-60|60-80|80-100>",
      "justification": "<...>",
      "context_violations": ["<list any context violations>"]
    }},
    "fluency": {{
      "score_100": <0-100>,
      "band": "<0-20|20-40|40-60|60-80|80-100>",
      "justification": "<...>"
    }},
    "coherence": {{
      "score_100": <0-100>,
      "band": "<0-20|20-40|40-60|60-80|80-100>",
      "justification": "<...>"
    }}
  }}
}}

RULES:
- Return ONLY valid JSON.
- Be strict. A score of 80+ means near-perfect. Penalize context violations heavily in naturalness.
\end{lstlisting}

\end{document}